\documentclass[letterpaper, 10 pt, journal,twoside]{IEEEtran}
\usepackage{amsmath,amsfonts}
\usepackage{algorithm}
\usepackage{algpseudocode}
\usepackage{algorithmicx}
\usepackage{mathabx}
\usepackage{array}
\usepackage[caption=false, font=footnotesize]{subfig}
\usepackage{array, makecell} %
\usepackage{textcomp}
\usepackage{stfloats}
\usepackage{url}
\usepackage{verbatim}
\usepackage{graphicx}
\usepackage{cite}

\usepackage{flushend}
\usepackage{graphicx}
\graphicspath{{images/}}
\usepackage{amsmath}
\usepackage{amssymb}
\usepackage{nicefrac}
\usepackage{amsfonts}
\usepackage{mathtools}
\usepackage{siunitx} 
\usepackage{textcomp}

\algrenewcommand\algorithmicindent{1em}%
\usepackage{comment}

\usepackage{nomencl}
\makenomenclature

\usepackage{pifont}

\usepackage{tablefootnote}
\usepackage{tabularx,booktabs}
\usepackage{multirow,makecell}
\usepackage[pagebackref=false,breaklinks=true,colorlinks=true,bookmarks=false]{hyperref}

\begin{document}
\bstctlcite{IEEEexample:BSTcontrol}

\title{Self-supervised Learning for Panoptic \\
Segmentation of Multiple Fruit Flower Species}

\author{Abubakar Siddique,~\IEEEmembership{Student Member,~IEEE,}
          Amy Tabb,~\IEEEmembership{Senior Member, IEEE} \\
          Henry Medeiros,~\IEEEmembership{Senior Member, IEEE}
\thanks{Abubakar Siddique is with the Department of Electrical and Computer Engineering, Marquette University, Milwaukee, USA, e-mail: abubakar.siddique@marquette.edu}%
\thanks{Amy Tabb is with the United States Department of Agriculture (USDA), Kearneysville, West Virginia, USA, e-mail: amy.tabb@usda.gov}%
\thanks{Henry Medeiros is with the Department of Agricultural and Biological Engineering, University of Florida, Gainesville, USA, e-mail: hmedeiros@ufl.edu} %
\thanks{Mention of a concept, idea, trade name, or commercial product in this publication is solely for the purpose of providing specific information and does not imply recommendation or endorsement by the US Department of Agriculture. The USDA is an equal opportunity employer. A. Tabb work was funded, in part, by USDA-ARS Project 8080-21000-032-00-D.}
\thanks{Manuscript received May 9, 2022.}}

\markboth{Robotics and Automation Letters,~Vol.~XX, No.~Y, August~2022}{Siddique \MakeLowercase{\textit{et al.}}: Self-supervised Learning for Panoptic Segmentation of Multiple Fruit Flower Species}


\pagestyle{empty}
\maketitle
\thispagestyle{empty}

\begin{abstract}
Convolutional neural networks trained using manually generated labels are commonly used for semantic or instance segmentation. In precision agriculture, automated flower detection methods use supervised models and post-processing techniques that may not perform consistently as the appearance of the flowers and the data acquisition conditions vary. We propose a self-supervised learning strategy to enhance the sensitivity of segmentation models to different flower species using automatically generated pseudo-labels. We employ a data augmentation and refinement approach to improve the accuracy of the model predictions. The augmented semantic predictions are then converted to panoptic pseudo-labels to iteratively train a multi-task model. The self-supervised model predictions can be refined with existing post-processing approaches to further improve their accuracy. An evaluation on a multi-species fruit tree flower dataset demonstrates that our method outperforms state-of-the-art models without computationally expensive post-processing steps, providing a new baseline for flower detection applications.
\end{abstract}


\section{Introduction}
Computer vision algorithms are becoming increasingly popular in agricultural applications. Detecting and counting flowers is an important crop management activity to optimize fruit production \cite{farjon2020detection}. Automatic bloom intensity estimation methods have the potential to reduce workloads in large production fields. Many machine vision approaches have been proposed to address the challenges of estimating crop yield. Most recent flower detection and counting methods based on deep learning models require a large amount of manually labeled training data to achieve acceptable performance \cite{multiSpecies_philipe_2018,MultiSpecies_semantic_2021, berries_uav_2020_CVPR_Workshops}. Although weakly supervised approaches \cite{apple_orchard_count_regression_2022} can simplify the training of convolutional neural networks (CNNs), they are not particularly effective to adapt large-scale, pre-trained models to unseen object categories.

Data augmentation \cite{test_aug_uncertainty_WANG2019,data_aug_train_text_2020} is a de facto standard technique to reduce the dependence on manual annotations when training deep neural networks. But in agricultural visual data, the appearance of objects of interest and the scene conditions vary significantly from one field to another. Besides, since agricultural production environments usually require images to be acquired from moving vehicles \cite{multiSpecies_philipe_2018,berries_uav_2020_CVPR_Workshops,uav_precision_agri}, the sun conditions and dense background clutter make this task challenging in terms of model generalization. Hence, we still need to generate enough manual labels for various species of crops to generalize the prediction models across species with significantly different appearance and backgrounds potentially comprised of semantically distinct elements.

Although deep CNNs can perform reasonably accurate pixel-level semantic predictions \cite{DBLP:SSL_semantics,multiSpecies_philipe_2018}, false alarms due to similarities between flowers, fruits at different stages of maturation, and  background objects limit potential opportunities for the application of computer vision algorithms to agricultural automation tasks. Instance \cite{panet_liu2018} and panoptic \cite{panoptic_FPN_2019} segmentation models might be able to better identify individual flowers or clusters of flowers and thus improve detection performance. 

To address the above challenges, inspired by the works presented in \cite{panoptic_FPN_2019,multicam_ssl_arxiv_2022,multiSpecies_philipe_2018}, we propose a novel self-supervised panoptic segmentation approach that leverages a small number of annotations for supervised learning (SL) and then adjusts the model to challenging unlabeled datasets. 
In summary, the main contributions of this work are:
\begin{itemize}
\item A robust self-supervised flower segmentation method that addresses typical agricultural visual data challenges in fruit tree orchards.
\item A novel panoptic pseudo-label generation technique for automatically updating the model for unlabeled datasets that contain severe clutter and illumination challenges.
\item A robust sliding-window-based training and testing approach that does not require additional post processing to refine the network predictions.
\item Extensive evaluations on multiple-species  datasets, which demonstrate superior generalized performance over state-of-the-art techniques.
\item Upon acceptance of this paper, our augmented datasets, code, and models will be made publicly available.
\end{itemize}




\begin{figure*}[t]
\centering

\includegraphics[width=0.99\textwidth]{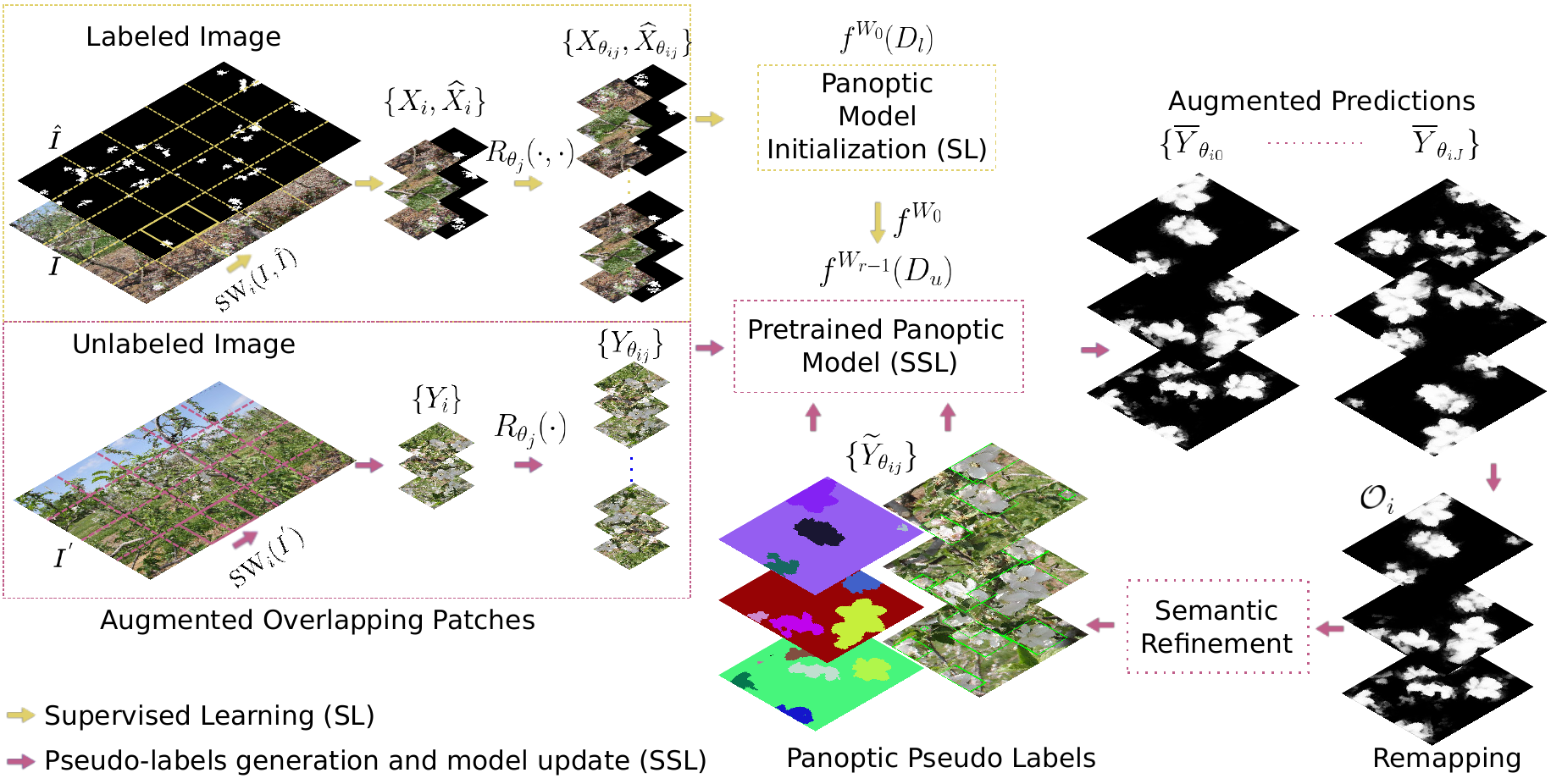}

\caption{Proposed self-supervised learning framework for multi-species flower segmentation.  Labeled images are used to initialize the model for flower segmentation. The overlapping sliding window patches of the unlabeled input images are rotated multiple times to generate the augmented semantic predictions from a previously initialized panoptic segmentation model. The remapping step transforms the score maps to the input coordinate system and then the normalized predictions are used to generate the panoptic pseudo-labels using 
a semantic refinement procedure to update the pre-trained model.}
\label{fig:model}
\end{figure*}

\section{Related Work}
In agricultural automation, several supervised \cite{chen2018encoder,bochkovskiy2020yolov4,he2017mask} and weakly supervised \cite{weak_superv_bmvc2019} deep learning models have been employed to address the challenges of detecting flowers \cite{multiSpecies_philipe_2018,MultiSpecies_semantic_2021,thinning_apple_flower_segmFCN2020,apple_flower_det_realYOLO2020}, fruits \cite{li2022real,fruits_det_alg_rtime_mango,berries_uav_2020_CVPR_Workshops},  or entire plants  \cite{plant_back_segm2020}. Applications of these methods range from robotic harvesting to estimating fruit load and optimizing fruit production by counting flowers in the early blooming season. Although some of these approaches leverage data augmentation techniques to generate automatic labels \cite{fruits_data_aug_frcnn2017,fruits_autolabel2020,multicam_ssl_arxiv_2022}, none of these methods addresses model generalization ability for significantly different test datasets. In the context of object detection and segmentation, recent methods attempt to accommodate data distribution shifts through the following techniques: a) supervised learning, b) semi-supervised learning, c) self-supervised learning, and d) multi-task panoptic segmentation models.

\paragraph{Supervised Methods}
These methods usually employ basic image transformations \cite{panet_liu2018, chen2018encoder} or sophisticated data augmentation techniques \cite{autoaug_2019,effective_data_aug_sl_2017} to improve model generalization. 
In addition to data augmentation during training, some methods incorporate post-processing algorithms at test time \cite{dias2018semantic,tang2021look}  or include specialized input/output units that are easier to fine-tune to new datasets \cite{point_rend_2020,crf_conv_BMVC_2020}. While these techniques reduce the dependency on annotations for different datasets, they do not eliminate it. Model performance is still largely dependent on the amount of training data available.

\paragraph{Semi-supervised Methods}
Using labeled data to bootstrap a model whose predictions are then employed to fine-tune the initial model (or to train a student model) is a popular approach to develop methods for multiple object detection \cite{SoftTeacher_2021_iccv}, as well as instance \cite{panet_liu2018,multicam_ssl_arxiv_2022} and semantic \cite{DBLP:SSL_semantics} segmentation. This strategy is effective when labeled and unlabeled data have similar appearance and sufficient labeled data is available to bootstrap a deep model. 
When the characteristics of the labeled and unlabeled data differ significantly, as is the case among different flower species, more sophisticated supervision mechanisms are needed  \cite{saito2020dance, ZSL2020semantics}.

\paragraph{Self-supervised Methods} When no labeled data is available, self-supervision strategies can be used to automatically generate pseudo-labels from the unlabeled data  \cite{SSL_dets_2019_CVPR, SSL_useful}.  In these scenarios, the initial model is trained to solve a surrogate task that presumably has a similar representation structure as the target task \cite{larsson2019fine}. Using unsupervised learning techniques to align latent feature representations is a widely used approach \cite{saito2020dance}. Self-supervision strategies that use model prediction uncertainties to guide the learning process, while arguably more interpretable and predictable, are less commonly explored. Our approach uses a multi-inference data augmentation mechanism in conjunction with the region growing refinement (RGR) algorithm \cite{dias2018semantic} to generate robust and accurate pseudo-labels in an iterative manner. These pseudo-labels allow our model to continuously improve its performance on previously unseen datasets.


\paragraph{Panoptic Methods}
Multi-task learning is commonly used to improve model performance across different tasks \cite{MTL2018Cipolla}. As long as the tasks are similar, the model tends to generalize better to unseen data \cite{multitask-segment}. The recently introduced panoptic segmentation approach jointly learns the closely related tasks of instance and semantic segmentation and currently represents the state of the art in instance and semantic segmentation \cite{mohan2020efficientps, li2022panoptic}. However, training such models requires a significant number of manual labels containing instance and semantic information. Our approach makes it possible to apply a panoptic model to  significantly different datasets without resorting to manual labels. To our knowledge no self-supervised panoptic segmentation method has been proposed so far.

\begin{figure*}
    \centering

     
     
     
    \subfloat[$\widebar{Y}_{\theta_{i0}}$]{\includegraphics[trim={0mm 0mm 0mm 105mm}, clip, width=0.25\linewidth]{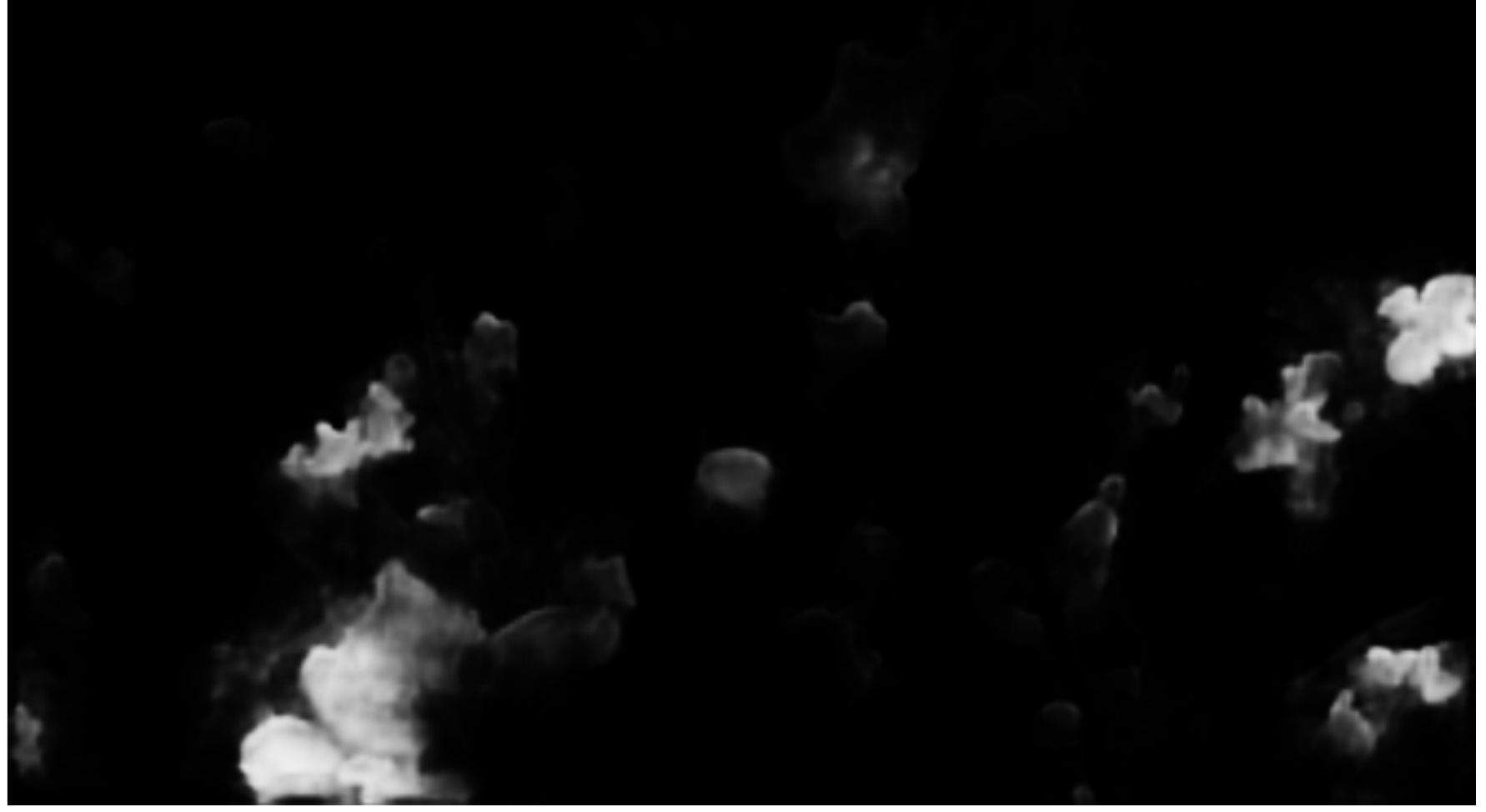}}
    \subfloat[${\mathcal{O}_i}$]{\includegraphics[trim={0mm 0mm 0mm 105mm}, clip,width=0.25\linewidth]{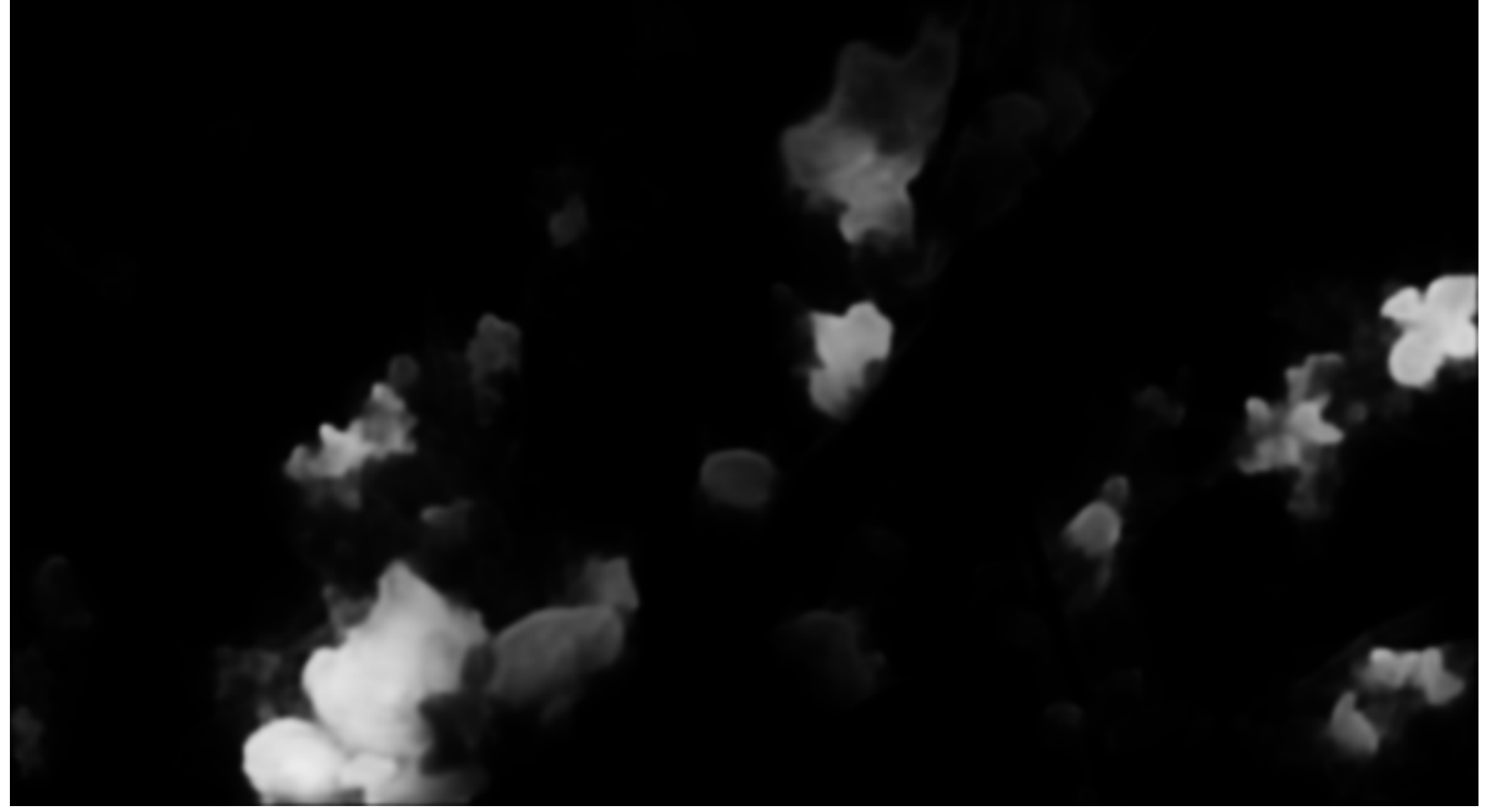}}
    \subfloat[$b_{\theta_{ij}}$]{\includegraphics[trim={0mm 0mm 0mm 105mm}, clip,width=0.25\linewidth]{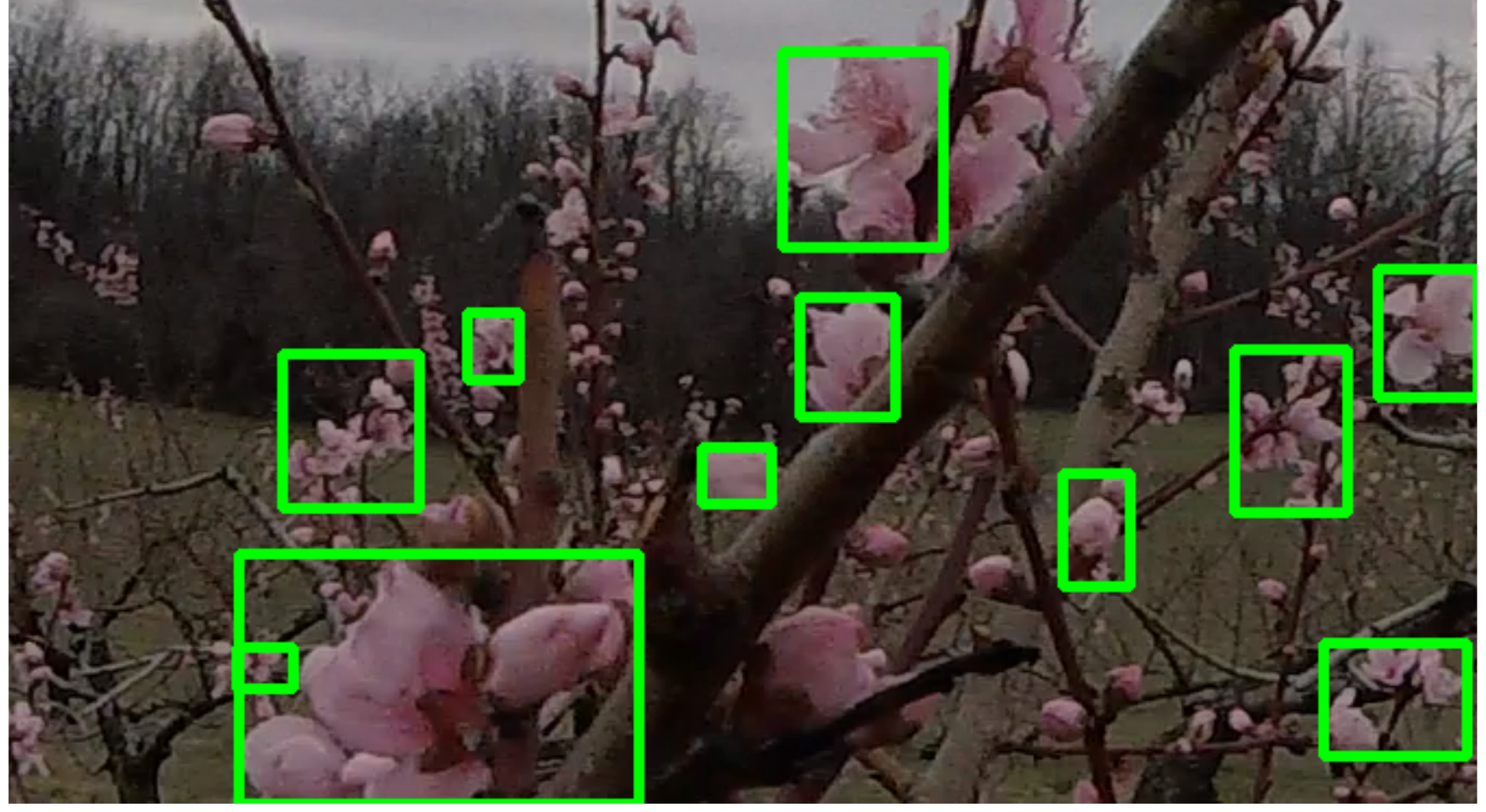}}
    \subfloat[$m_{\theta_{ij}}, S_{\theta_{ij}}$]{\includegraphics[trim={0mm 0mm 0mm 105mm}, clip,width=0.25\linewidth]{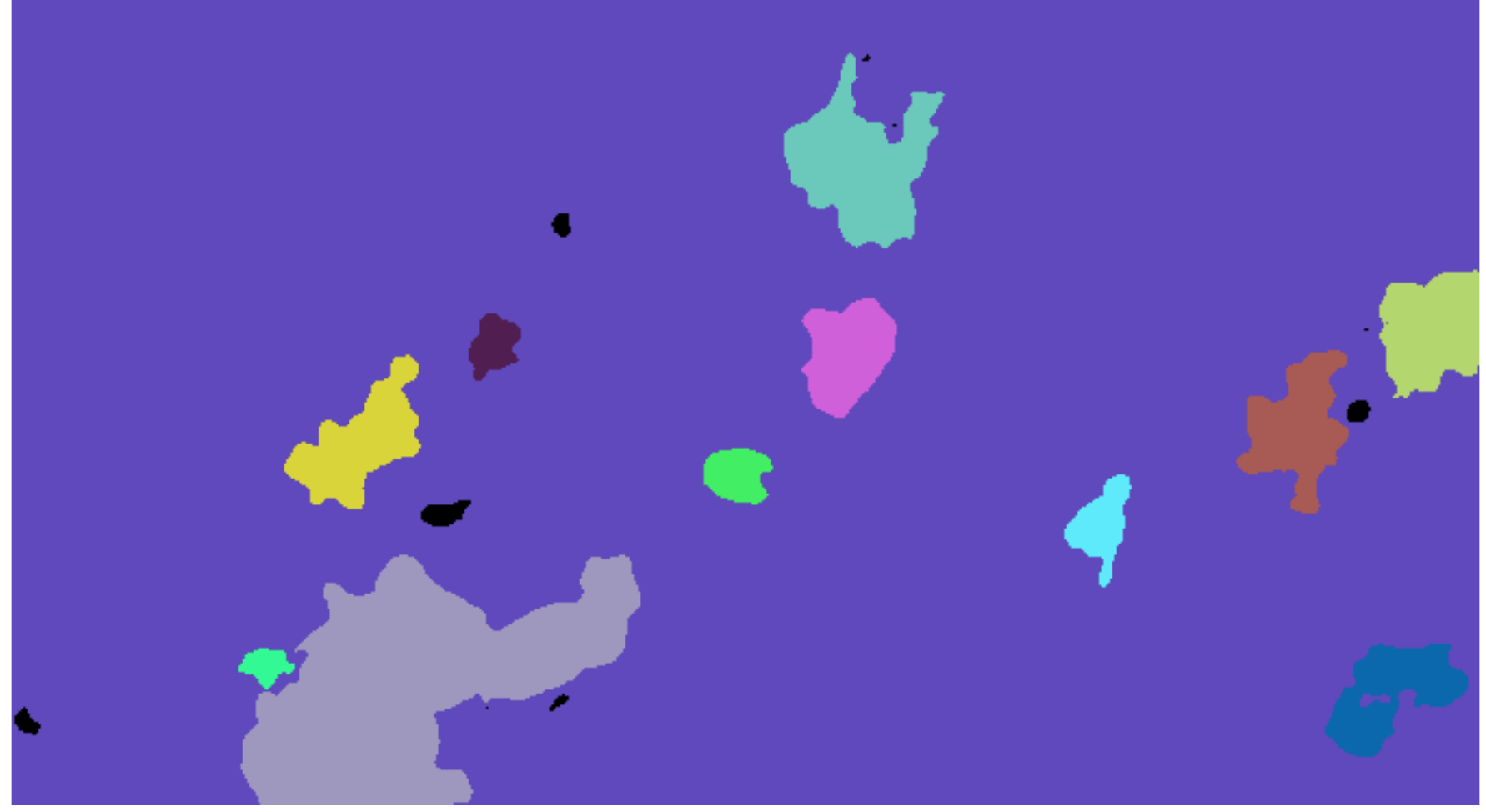}}
    
\caption{Illustration of the steps of our panoptic pseudo-label generation method. a) semantic prediction for a single augmented patch, b) normalized average scoremap obtained using Eq. \ref{eq:soft_scores},  c) instance bounding boxes, and d) instance segmentation masks and semantic labels generated during SSL iterations.}
\label{fig:quali_ablation}
\end{figure*}

\section{Self-supervised Panoptic Segmentation}
Our proposed self-supervised learning (SSL) technique for panoptic segmentation  shown in Fig. \ref{fig:model} comprises three main components: i) labeled and unlabeled data augmentation, ii) panoptic model initialization using the labeled dataset, and iii) panoptic pseudo-label generation from unlabelled data to update the model. As shown in Alg. \ref{alg:ssl_flower}, we use images from the training set and their corresponding labels to train our initial model using an SL strategy. Our SSL  approach then updates the initial model iteratively in a fully self-supervised manner using the pseudo-labels generated by the model at a previous iteration.

\begin{algorithm}[b]
\caption{Self-supervised Learning Algorithm} 
\label{alg:ssl_flower}
\begin{algorithmic}[1]
\Require{Set of high resolution labeled images $I$, their corresponding segmentation labels $\hat{I}$, and the set of unlabelled images $I^{'}$}
\Ensure{Self-supervised model $f^{W_r}$ for unlabeled data $I^{'}$}

    \State \parbox[t]{\dimexpr\linewidth-\algorithmicindent} {Generate the augmented training set $D_l$ using $I$ and $\hat{I}$ according to Eq. \ref{eq:data_aug}}
    \State \parbox[t]{\dimexpr\linewidth-\algorithmicindent} {Train the initial model $f^{W_0}(D_l)$ using $D_l$}
    \State \parbox[t]{\dimexpr\linewidth-\algorithmicindent} {Generate the augmented unlabelled image patches ${Y}_{\theta_{ij}}$}
    \For {$r \gets 1$ to $\text{maxIter}$}
        \State \parbox[t]{\dimexpr\linewidth-\algorithmicindent} {Generate the augmented predictions $\overline{Y}_{\theta_{ij}}$ using Eq. \ref{eq:class_logits}}
        \State \parbox[t]{\dimexpr\linewidth-\algorithmicindent} {Compute the normalized score map $\mathcal{O}_i$ using Eq. \ref{eq:soft_scores}}
        \State \parbox[t]{\dimexpr\linewidth-\algorithmicindent} {Compute the binary semantic mask $S_i$ from $\mathcal{O}_i$ using RGR \vspace{2pt}}
        \State \parbox[t]{\dimexpr\linewidth-\algorithmicindent} {Generate the augmented binary semantic masks $S_{\theta_{ij}}$}
        \State \parbox[t]{\dimexpr\linewidth-\algorithmicindent} {Apply connected component analysis to $S_{\theta_{ij}}$ to find the instance masks $m^{(l)}_{\theta_{ij}}$ and bounding boxes $b^{(l)}_{\theta_{ij}}$}
        \State \parbox[t]{\dimexpr\linewidth-\algorithmicindent-3pt} {Construct the set of pseudo-labels $\widetilde{Y}_{\theta_{ij}}$ using Eq. \ref{eq:pseudo-labels}}
        \State \parbox[t]{\dimexpr\linewidth-\algorithmicindent} {Construct the set $D_u=\{Y_{\theta_{ij}},\widetilde{Y}_{\theta_{ij}}\}$}
        \State \parbox[t]{\dimexpr\linewidth-\algorithmicindent} {Update the self-supervised model $f^{W_{r-1}}(D_u)$ using $D_u$}
    \EndFor

\end{algorithmic}
\end{algorithm}

\subsection{Data Augmentation}
\label{subsec:data_aug}
Our method is based on the panoptic segmentation model proposed in \cite{panoptic_FPN_2019} pre-trained on the COCO \cite{lin2014microsoft} and COCO-stuff \cite{caesar2018coco} datasets. To fine-tune the model for flower segmentation, we augment the training set introduced in \cite{multiSpecies_philipe_2018} using a sliding window ($\text{SW}$) technique. That is, we extract from the input image $I$ and its corresponding semantic label $\hat{I}$, both of size $M\times N$ pixels,  overlapping patches of size $m \times n=\lfloor \nicefrac{M}{K} \rfloor \times \lfloor \nicefrac{N}{K} \rfloor$ pixels with a stride of $p\times q= \lceil \nicefrac{m}{2} \rceil \times \lceil \nicefrac{n}{2} \rceil$, where $K$ is the window size factor. Let $(X_i,\widehat{X}_i)=\text{SW}_i(I,\hat{I})$ be the $i$-th image patch and its corresponding semantic label. We augment $X_i$ and $\hat{X}_i$ by applying  $J$ different rotations at randomly selected angles $\{\theta_j\}_{j=0}^J$. For the sake of sampling efficiency, rather than directly sampling from the interval $[0,2\pi]$, we employ a stratified sampling strategy. That is, we partition the circle into five sectors centered at $\left(\nicefrac{\pi}{2}\right)\cdot k$, $k=0,1,\ldots,4$ and sample each sector uniformly. This strategy increases sample diversity, ultimately reducing the variance of the pseudo-labels generated using our method.
Thus, the set of labeled image patches and corresponding manual labels used to train the supervised model is given by
\begin{equation}
 D_{l}=\left\{\left(X_{\theta_{ij}}, \widehat{X}_{\theta_{ij}}\right)\right\}=\left\{R_{\theta_j}(\text{SW}_i(I, \hat{I}))\right\}, 
 \label{eq:data_aug}
\end{equation}
where $R_{\theta_j}(\cdot,\cdot)$ rotates its two arguments by an angle $\theta_j$.

We employ the same data augmentation procedure for each unlabeled image of the test sets to generate the unlabeled augmented samples $Y_{\theta_{ij}}$ from the corresponding image patches $Y_i$. In the SSL approach, we use the SL model to predict the initial augmented pseudo-labels $\widetilde{Y}_{\theta_{ij}}$ used to fine-tune the model for unseen datasets. The procedure for pseudo-label generation is described in detail in Section \ref{sec:pseudo-label}.
Thus the unlabeled dataset for each flower species is 
\begin{equation}
D_{u}=\left\{\left(Y_{\theta_{ij}}, \widetilde{Y}_{\theta_{ij}}\right)\right\}.
\label{eq:data_aug_unlab}
\end{equation}

At test time, we simply apply the sliding window operation to generate the normalized semantic score maps and combine the predictions corresponding to the overlapping portions of each window using majority voting. We  observed that the benefit of test-time data augmentation is negligible after a few SSL training iterations. Hence, we do not perform rotation augmentation at inference time, which ensures that the computational time of the model remains unchanged. 


\subsection{Pseudo-label Generation}
\label{sec:pseudo-label}
Data distribution shifts degrade the accuracy of segmentation models. Strong data augmentation is an effective strategy to mitigate this problem \cite{Yuan_2021_ICCV}. Thus, to improve the sensitivity of our model to different flower species, we apply the data augmentation procedure described above to $Y_i$ and use the previously computed network weights $W_{(r-1)}$ to generate the augmented predictions at the $r$-th SSL iteration according to
\begin{equation}
\overline{Y}_{\theta_{ij}}= f^{W(r-1)}(Y_{\theta_{ij}}).
\label{eq:class_logits}
\end{equation}
To remap the semantic predictions back to the original image coordinate frame, we apply the reverse rotation operator $R_{-\theta_{j}}(\cdot)$ with bi-linear interpolation to the augmented predictions $\overline{Y}_{\theta_{ij}}$. We then normalize the scores using a softmax function and use the average normalized score map $\mathcal{O}_i$ as our  final prediction, i.e.,
\begin{equation}
\mathcal{O}_i=\frac{1}{J}\sum_{j} \sigma\left(R_{-\theta_{j}}(\overline{Y}_{\theta_{ij}})\right),
\label{eq:soft_scores}
\end{equation}
where $\sigma(\cdot)$ represents the softmax function applied element-wise to the individual logits for the classes $\mathcal{C} \in \left\{\mathrm{background}, \mathrm{flower}\right\}$. As  Figs. \ref{fig:quali_ablation} (a) and (b) illustrate, $\mathcal{O}_i=$ contains a significantly higher number of flowers segmented with high confidence than a single augmented patch $\overline{Y}_{\theta_{ij}}$.

\begin{figure*}[t]
\includegraphics[width=0.99\linewidth]{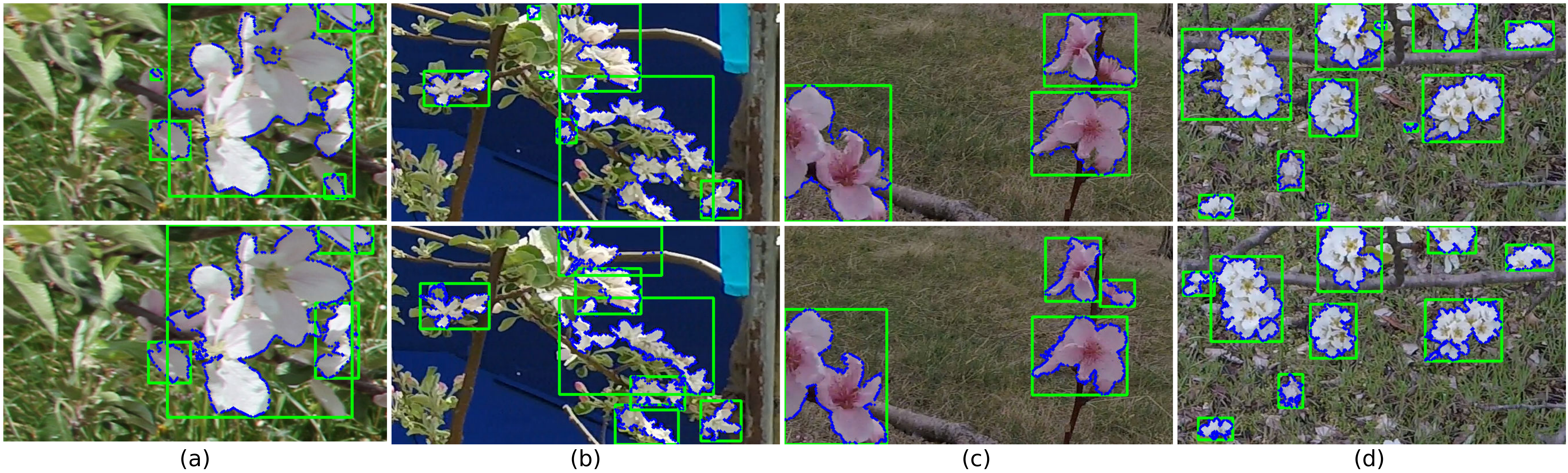}
\centering
\caption{Comparisons between the pseudo-labels generated using  a fixed threshold $\tau_{seg}$ (top row) and the RGR-based semantic refinement (bottom row). a) AppleA, b) AppleB, c) Peach, d) Pear. The segmentation masks in the images at the bottom row better reflect flower boundaries and the corresponding bounding boxes better distinguish nearby flower instances.}
\label{fig:semantic_refine}
\end{figure*}

\subsection{Semantic Prediction Refinement}
\label{sec:sem_seg_refine}
Instead of applying a hard threshold
to generate panoptic pseudo-labels from $\mathcal{O}_i$, we employ RGR, a robust segmentation refinement method  \cite{dias2018semantic}. RGR uses a Monte Carlo strategy to perform an appearance-based refinement of low-confidence regions in $\mathcal{O}_i$ using the corresponding image patch $Y_i$, which allows it to generate an improved binary segmentation mask. RGR uses three key elements to determine the boundaries of an object of interest: 1) the confidence of the model predictions, 2) appearance similarities among pixels, and 3) distances among pixels. That is, every pixel in an image is associated with a nearby pixel of similar appearance whose semantic class has been predicted with high confidence. As Fig. \ref{fig:semantic_refine} illustrates, RGR improves the boundary adherence of the pseudo-labels and better distinguishes flower instances.

Let $S_i$ be the semantic binary mask obtained from $\mathcal{O}_i$ using RGR. As in the pseudo-label generation step, we apply $J$ rotations to $S_i$ to generate augmented semantic binary masks, $S_{\theta_{ij}}=R_{\theta_{j}}(S_i)$. We then perform connected component analysis to obtain the  corresponding instance masks $m^{(l)}_{\theta_{ij}}$ and bounding boxes $b^{(l)}_{\theta_{ij}}$ for the $l=1,\ldots,L$ distinct elements of $S_{\theta_{ij}}$. The augmented panoptic pseudo-labels are given by
\begin{equation}
\widetilde{Y}_{\theta_{ij}}=\left\{(b^{(l)}_{\theta_{ij}}, m^{(l)}_{\theta_{ij}}), S_{\theta_{ij}} \right\}_{l=1}^{L}.
\label{eq:pseudo-labels}
\end{equation}
Figs. \ref{fig:quali_ablation} (c) and (d) show that this approach generates high-quality bounding boxes and instance masks.

\subsection{Multi-task Loss}

In both the SL and SSL models, the instance bounding boxes $b^{(l)}_{\theta_{ij}}$ and segmentation masks $m^{(l)}_{\theta_{ij}}$ from the augmented labels are used to train the ROI-heads for the $\mathrm{flower}$ class. The augmented semantic masks $S_{\theta_{ij}}$ are used to train the semantic segmentation head for the $\mathrm{background}$ and $\mathrm{flower}$ classes. For panoptic segmentation learning, we consider $\mathrm{background}$ as a stuff class and $\mathrm{flower}$ as a thing class \cite{kirillov2019panoptic} to jointly update the model using the following multi-task loss function
\begin{equation}
    \mathcal{L}(W) = \lambda (\mathcal{L}_c+\mathcal{L}_b+\mathcal{L}_m) + (1-\lambda)\mathcal{L}_s,
    \label{eq:loss_panoptic}
\end{equation}
where $\mathcal{L}_c$ is the classification loss, $\mathcal{L}_b$ is the bounding-box loss, $\mathcal{L}_m$ is the mask loss, and $\mathcal{L}_s$ is the segmentation loss, as defined in \cite{panoptic_FPN_2019}.
By further training the initial SL model on the unlabeled datasets using the proposed SSL approach where the augmented panoptic labels  are robust to prediction uncertainty and intrinsically incorporate rotation invariance, it is possible to iteratively improve the performance of the model.

\section{Experiments}

We compare the performance of our method against the state-of-the-art algorithms presented in \cite{multiSpecies_philipe_2018,MultiSpecies_semantic_2021} using the evaluation metrics and procedures described in \cite{multiSpecies_philipe_2018}. To quantify the benefit of employing RGR as part of our pseudo-label generation strategy, we evaluate two different techniques to generate the pseudo-labels. First, we evaluate an approach in which we apply a hard threshold $\tau_{seg}$ to the predicted scoremaps. For a fair comparison, we determine $\tau_{seg}$ based on the maximum $F_1$ score obtained by the model on the training set at a previous iteration (see Fig. \ref{fig:PR_curve}). We call this model \texttt{SSL}. The model in which we employ RGR to refine the scoremaps without hard thresholding is deemed \texttt{SSL+RGR}. We also assess the performance improvements obtained by applying RGR as a post-processing mechanism in conjunction with our SSL model. We refer to that approach as \texttt{SSL+RGR~(pp)}, where \emph{pp} stands for post-processing. As a baseline, we also assess the performance of the \texttt{SL} model trained only on the AppleA dataset applied to the other datasets.

\subsection{Datasets}
We evaluate our method on the multi-species flower dataset first introduced in \cite{multiSpecies_philipe_2018}, which comprises four subsets: i) AppleA (train/test), ii) AppleB, iii) Peach, and iv) Pear. We train our SL model using the AppleA training set, which consists of $100$ images with a resolution of $M \times N=5184 \times 3456$ \cite{multiSpecies_philipe_2018}. After applying $J$ rotation augmentation steps, the number of training patches $X_{\theta_{ij}}$ for each input image is $J \times (2K-1)^2$ since $i=1,2,\ldots,(2K-1)\times(2K-1)$ and $j=1,2,\ldots,J$.  Hence, for $K=4$ and $J=20$, there are $98,000$ training patches in the AppleA dataset. These patches are used to train our initial panoptic flower segmentation model. 

We consider a randomly selected subset comprising $70\%$ of the $30$ images from the AppleA test set 
as unlabeled images $I^{'}$ to fine-tune the SL model using the automatically generated panoptic pseudo-labels. Similarly, $70\%$ of the images from the AppleB, Peach, and Pear datasets ($18$, $24$, and $18$ images, respectively), all of which have a resolution of $2704\times 1520$, are considered unlabeled images used to update the SL model iteratively. The remaining images in each dataset are used solely for performance evaluation. Given the relatively small size of the test sets, we evaluate our methods using five-fold cross-validation. 
\begin{figure}[h]
\includegraphics[width=0.99\linewidth]{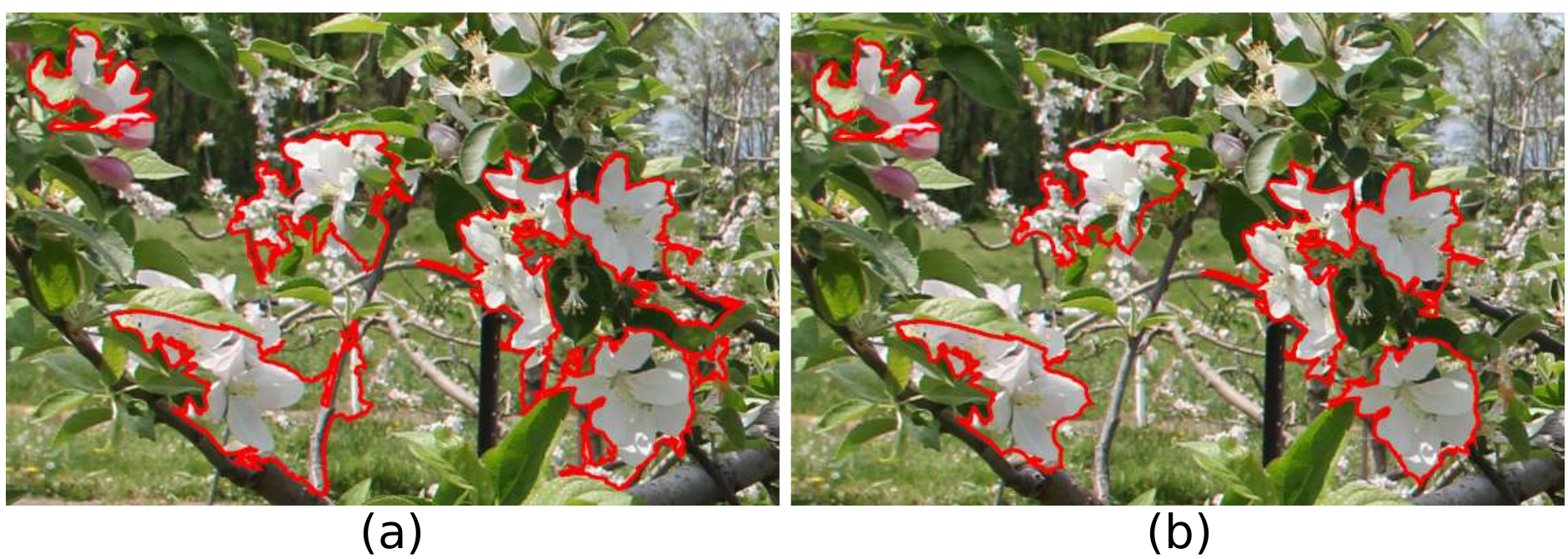}
\centering
\caption{Examples of improved annotations in the AppleA training set. The cropped sections shows (a) incorrect contours containing background pixels, and (b) improved labels.}
\label{fig:gt_correction}
\end{figure}


The datasets introduced in \cite{multiSpecies_philipe_2018} provide pixel-level, high-resolution annotations of individual flowers. However, as Fig. \ref{fig:gt_correction} shows, the annotations have imperfections that can only be observed when closely inspected. Despite being small, these inaccuracies comprise a non-negligible portion of the image pixels, especially considering that only a fraction of the pixels correspond to flowers.  To resolve this issue, we use the MATLAB$^{\circledR}$ image labeler tool to manually correct inaccurate labels and to label additional smaller but clearly visible unannotated flowers. Fig. \ref{fig:gt_correction} shows some examples of the annotations before and after the corrections.


\subsection{Training Details}
\label{subsec:training_details}
The vast majority of image pixels in the datasets correspond to background pixels. Hence, to provide the model sufficient samples containing flower pixels, we train the network for $20,000$ iterations using stochastic gradient descent with a batch size of $512$ samples and a base learning rate of $25\mathrm{e}{-4}$, which is divided by $10$ at $10\%$, $25\%$, and $50\%$ of the training period. We freeze the ResNet-101 backbone \cite{he2016deep} during training. To emphasize semantic learning, we use $\lambda=0.8$ in Eq. \ref{eq:loss_panoptic}.  
We have empirically observed that setting RGR's average spacing between samples to $100$ pixels provides an adequate balance between the accuracy of the refined scoremap and the computation required to produce it. We use the values reported in \cite{multiSpecies_philipe_2018} for the remaining parameters, namely, the number of iterations is $10$, the scoremap threshold is $0.5$, the high-confidence foreground threshold is $0.8$, and the high-confidence background threshold is $0.01$. 


\section{Results and Discussion}

\begin{table}[h]
\setlength{\tabcolsep}{3pt}
\caption{Evaluation of flower segmentation performance using our SSL panoptic model. The best results are shown in boldface and the second-best are underlined. We report the average value of the evaluation measures and their standard deviations across five runs.}
\centering
\label{tab:flower_seg_eval_pan}
\begin{tabular}{llllll}
\toprule
Dataset                 & Method              & IoU & F1 & Rcll & Prcn \\
\toprule
\multirow{6}{*}{AppleA} & DeepLab+RGR \cite{multiSpecies_philipe_2018}              &71.4     &83.3    &87.7      &79.4      \\
                        & DeepLab+SCL \cite{MultiSpecies_semantic_2021}  &\bf{81.1}     &\bf{89.6}    &\bf{91.9}      &87.3      \\
                        & SL      &77.1$\pm$0.9     &87.0$\pm$0.5   &86.7$\pm$0.6      &87.3$\pm$0.8      \\
                        & SSL      &76.2$\pm$0.6     &86.1$\pm$0.7    &88.2$\pm$0.9      &84.8$\pm$0.9      \\

                        & SSL+RGR &77.9$\pm$0.6     &87.5$\pm$0.3    &87.8$\pm$0.6      &\underline{87.3$\pm$0.6}      \\
                        & SSL+RGR (pp)      &\underline{79.6$\pm$0.6}     &\underline{88.6$\pm$0.3}    &\underline{89.2$\pm$0.6}      &\bf{88.1$\pm$0.7}      \\
\hline
\multirow{6}{*}{AppleB} & DeepLab+RGR \cite{multiSpecies_philipe_2018}              &63.0     &77.3    &\bf{91.2}      &67.1      \\
                        & DeepLab+SCL \cite{MultiSpecies_semantic_2021}   &65.3     &79.6    &72.7      &87.4      \\
                        & SL      &75.8$\pm$0.8     &86.2$\pm$0.5    &85.4$\pm$1.1      &87.1$\pm$0.5      \\
                        & SSL     &76.8$\pm$0.7     &86.8$\pm$0.4    &87.0$\pm$0.7      &86.7$\pm$0.8      \\
                         & SSL+RGR &\underline{78.7$\pm$0.4}     &\underline{88.1$\pm$0.2}    &\underline{87.9$\pm$0.3}      &\underline{88.2$\pm$0.7}      \\
                        & SSL+RGR (pp)      &\bf{79.9}$\pm$0.8  &\bf{88.9}$\pm$0.5    &86.7$\pm$1.0      &\bf{92.2}$\pm$0.3      \\
\hline
\multirow{6}{*}{Peach}  & DeepLab+RGR \cite{multiSpecies_philipe_2018}              &59.0     &74.2    &64.8      &86.8      \\
                        & DeepLab+SCL \cite{MultiSpecies_semantic_2021}  &64.3    &77.7      &70.3   &\underline{88.2}   \\
                        & SL      &48.9$\pm$3.5     &65.6$\pm$3.2    &62.6$\pm$4.2      &68.9$\pm$2.6      \\
                        & SSL   &67.8$\pm$4.1    &80.7$\pm$2.9    &\bf{85.3}$\pm$2.1 &76.7$\pm$3.6     \\
                        & SSL+RGR &\underline{75.2$\pm$3.2}     &\underline{85.8$\pm$2.1}    &\underline{84.6$\pm$1.9}      &86.9$\pm$2.4      \\
                        & SSL+RGR (pp)      &\bf{78.3}$\pm$3.2     & \bf{87.8}$\pm$1.7   &84.9$\pm$2.1     &\bf{91.1}$\pm$3.0      \\
\hline
\multirow{6}{*}{Pear}   & DeepLab+RGR \cite{multiSpecies_philipe_2018}               &75.4     &86.0    &79.2      &94.1      \\
                        & DeepLab+SCL \cite{MultiSpecies_semantic_2021}   &74.5     &85.4    &75.4      &\bf{97.3}      \\
                        & SL      &77.3$\pm$1.9     &87.2$\pm$1.3    &85.1$\pm$2.4      &89.4$\pm$0.7      \\
                        & SSL   &78.6$\pm$1.7     &87.9$\pm$1.0    &\underline{87.9$\pm$1.6}      &88.1$\pm$0.8     \\
                        & SSL+RGR      &\underline{82.4$\pm$1.9}     &\underline{90.4$\pm$1.2 }   &\bf{89.4}$\pm$1.8      &91.3$\pm$1.4      \\
                        & SSL+RGR (pp)      &\bf{84.2}$\pm$2.1     &\bf{91.4}$\pm$1.2    &87.4$\pm$1.9      &\underline{95.8$\pm$0.9}      \\
\bottomrule
\end{tabular}
\end{table}

\begin{figure}[t]
\includegraphics[width=0.99\linewidth]{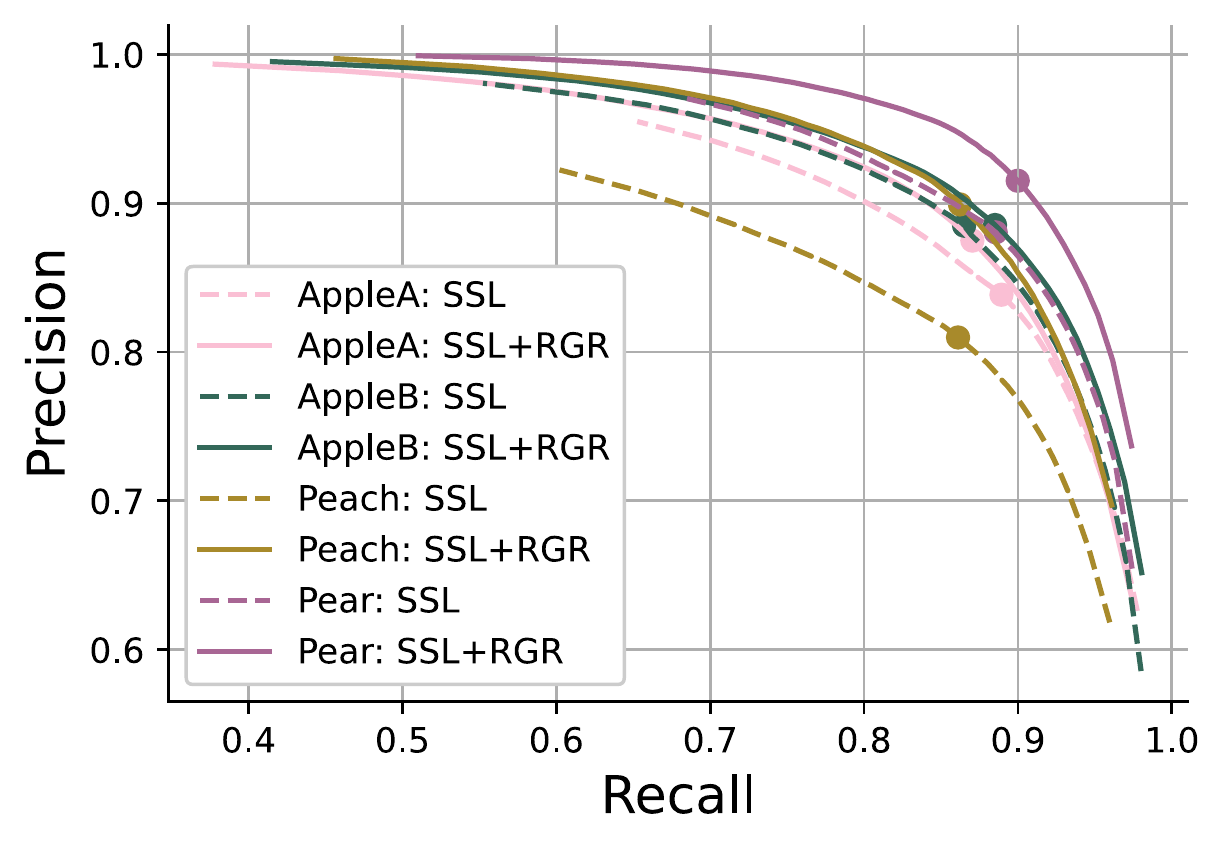}
\centering
\caption{Precision-recall curves for the SSL models with and without RGR pseudo-label refinement. Solid circles represent points that maximize $F_1$ scores.}
\label{fig:PR_curve}
\end{figure}

\begin{figure*}[h]
\includegraphics[width=0.99\linewidth]{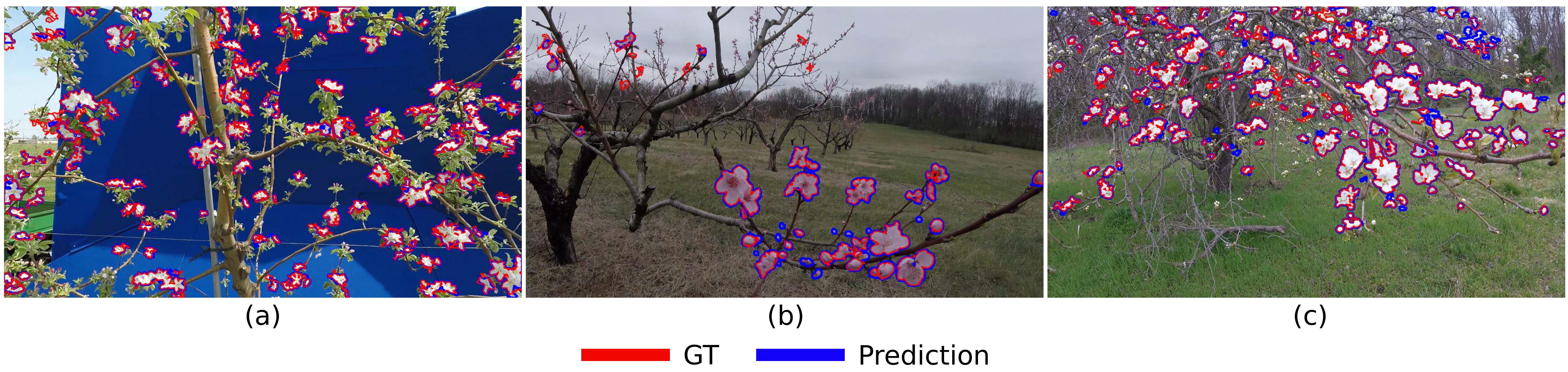}
\centering
\caption{Qualitative assessment of our proposed SSL approach on test datasets (a) AppleB, (b) Peach, (c) Pear. Most false positives are unlabeled flowers.}
\label{fig:qauli_full}
\end{figure*}
Table \ref{tab:flower_seg_eval_pan} compares the performance of the SL and SSL models against the algorithms presented in \cite{multiSpecies_philipe_2018,MultiSpecies_semantic_2021}. Although the SL model trained using our proposed data augmentation strategy segments flowers using a fixed threshold $\tau_{seg}$, it performs either on par with or better than the state-of-the-art models on test sets that are similar to the training set, even without applying our proposed SSL strategy. However, for datasets with significantly different characteristics, the SL model does not perform satisfactorily.  The SSL approach using a hard threshold outperforms the baseline methods on the AppleB, Peach, and Pear datasets by significant margins ($11.5\%$, $3.5\%$, and $4.1\%$ absolute IoU improvement with respect to \cite{MultiSpecies_semantic_2021}). For the AppleA dataset, the SSL method alone outperforms \cite{multiSpecies_philipe_2018} but is slightly worse than \cite{MultiSpecies_semantic_2021}. This is largely due to the fact that the baseline methods perform dramatically better on the training set, whereas the performance of our model remains relatively stable across datasets. As discussed in more detail below, background flowers also contribute to the performance degradation. When we use RGR to refine the pseudo-labels, we observe an IoU improvement of up to $7.4\%$ with respect to the SSL method using a fixed threshold. Performing an additional RGR step at test time leads to an additional average IoU improvement of approximately $1.9\%$ but at the cost of substantially higher inference times, as discussed in the next section. Fig. \ref{fig:PR_curve} shows the precision-recall curves for the proposed SSL methods with and without pseudo-label refinement using RGR. 

The qualitative results in Fig. \ref{fig:qauli_full} show that the SSL models are highly sensitive to flowers in complex regions. For some datasets, the SSL methods show slightly lower precision than \cite{MultiSpecies_semantic_2021}. The main reason for the lower precision is the presence of small, unannotated flowers in the datasets that our model can detect. This can be observed in Fig. \ref{fig:qauli_full} (c) where several small flowers are present, especially on branches farther from the camera. Determining which flowers should be annotated is an application-specific problem that requires further investigation.

\subsection{Parameter Sensitivity and Computation Time Analysis}
\begin{figure}[h]
\includegraphics[width=0.99\linewidth]{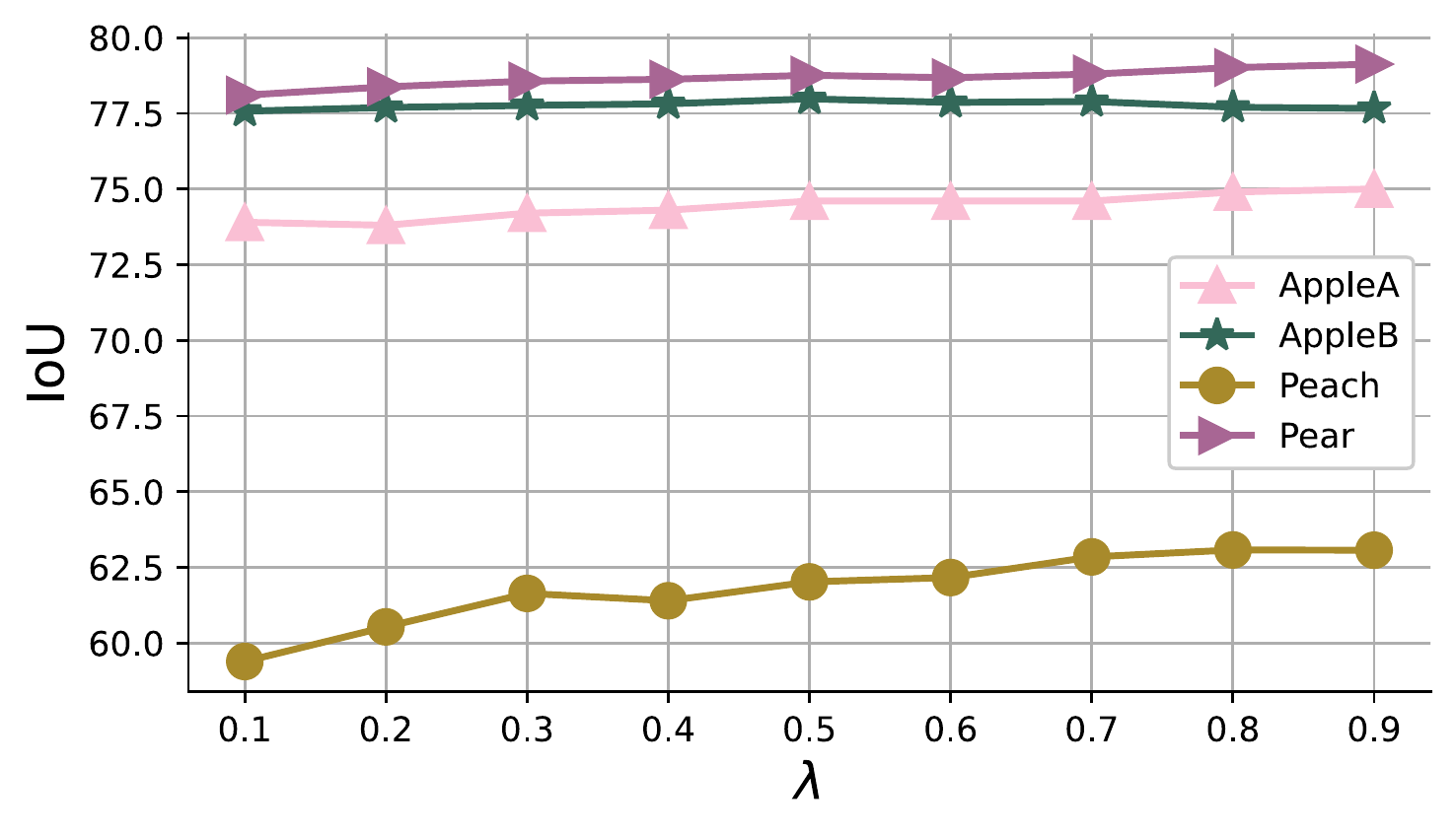}
\centering
\caption{Impact of the loss weight $\lambda$ (Eq. \ref{eq:loss_panoptic}) on flower segmentation performance at the first SSL iteration with $J=20$ and $K=4$.}
\label{fig:lambda_ablation}
\end{figure}

Table \ref{tab:ablation_sw} shows the impact of the sliding window size factor $K$ and the number of rotation angles $J$ on model performance and average inference time per input image. This evaluation is performed on the first SSL iteration of a model initialized with $K=4$ and $J=20$. That is, the evaluation reflects the impact of model parameters on the accuracy of the resulting pseudo-labels. The top two rows show the test-time impact of varying $K$ without employing test-time rotations (i.e., $J=1$) for the AppleA and AppleB datasets, respectively. The last row of the table shows that the IoU and $F_1$ measures on the Peach dataset gradually increase with $J$ when rotation augmentation is employed at inference time,  but so does the computation time. Inference times were obtained using one NVIDIA$^{\circledR}$ GeForce$^{\circledR}$ RTX 2080 Ti GPU without any multi-processing technique. Post-processing times using RGR are approximately $16\times$ higher than those presented in Table \ref{tab:ablation_sw} on our Intel$^{\circledR}$
 Xeon$^{\circledR}$
 Silver 4112 CPU $@ 2.6G$Hz. Results for the remaining datasets are similar and are omitted for brevity. Fig. \ref{fig:lambda_ablation} shows the impact of $\lambda$ in the multi-task loss (Eq. \ref{eq:loss_panoptic}) for different flower species. Although the performance of our approach remains relatively stable as we vary $\lambda$, for most datasets, the best results are obtained with $0.7\leq\lambda\leq0.9$, especially in cross-species scenarios, where appearance variation is more prominent. 
\begin{table}[t]
\caption{Performance impact of sliding window size and number of rotation angles.}
\centering
\label{tab:ablation_sw}
\setlength\tabcolsep{6pt}
\begin{tabular}{ccccccc}
\toprule
Dataset & $M \times N$                        &$K$     &$J$  & IoU & $F_1$  & \makecell{Inf. Time \\  (sec.)}\\
\hline

\multirow{3}{*}{AppleA} & \multirow{3}{*}{$5184 \times 3456$} &4     & \multirow{3}{*}{$1$}   &73.6    &84.8  &7.2 \\
                                    & &8     &                         &75.4    &86.0  &15.4 \\
                                    & &16    &                         &53.3    &69.5  &90.0 \\
                           \hline
                           
\multirow{3}{*}{AppleB} & \multirow{3}{*}{$2704 \times 1520$} &2    & \multirow{3}{*}{$1$} &71.6    &83.0  &1.4 \\
                                    & &4    &                       &76.7   &86.8  &5.3 \\ 
                                    & &8    &                       &57.1   &72.6  &22.1 \\
\hline
 \multirow{4}{*}{Peach} & \multirow{4}{*}{$2704 \times 1520$} & \multirow{4}{*}{$4$} &1  &51.3  &67.8     &5.5  \\
                                    &        &                       &5         &58.2  &73.6     &34.7         \\
                                    &        &                       &10         &60.3     &75.2      &71.5   \\
                                    &        &                       &20      &61.3     &76.0    &147.8 \\
\bottomrule
\end{tabular}
\end{table}

\section{Conclusions}
We introduce a self-supervised learning technique to accurately segment multiple tree flower species without significant manual labeling efforts. To automatically generate instance and semantic labels for unlabeled datasets, we propose a data augmentation technique associated with a semantic segmentation refinement strategy that produces accurate pseudo-labels for self-supervised model training. The proposed SSL technique makes it possible to train a deep multi-task model effectively on unlabeled fruit flower datasets. Self-supervised learning substantially reduces model dependency on computationally expensive post-processing steps to further refine the model predictions at inference time. That being said, employing a post-processing approach with our SSL model can further improve its prediction accuracy. Our novel SSL method creates a new baseline for the multi-species flower segmentation task. 

In the future, panoptic flower segmentation can be further improved in a number of ways. First, our proposed framework resorts primarily to a data augmentation strategy based on image rotations. Given the characteristics of the problem under consideration, it stands to reason that additional data augmentation strategies such as color jittering and image blurring would further contribute to the generation of accurate pseudo-labels. In addition, instead of using empirically defined weights for the instance and semantic segmentation tasks, task-dependent uncertainty learning strategies \cite{siddique2021unsupervised} may better capture appearance variations to optimize the task weights. Finally,  pseudo-label pixels or sometimes entire instances may have low prediction scores. The uncertainty of the pseudo-labels may be used to weigh the contributions of individual samples. Uncertainty-weighed loss functions \cite{multicam_ssl_arxiv_2022} are a promising technique to accomplish that goal. 

\bibliographystyle{IEEEtran}
\bibliography{IEEEabrv,egbib}

\end{document}